\documentclass[final,3p]{elsarticle}

\usepackage{lineno}
\usepackage{amssymb}  
\usepackage{amsmath}  
\usepackage[colorlinks=true,breaklinks=true,pdftex]{hyperref}
\modulolinenumbers[1]
\usepackage{booktabs} 
\usepackage{multirow}
\usepackage[normalem]{ulem}
\usepackage{amssymb}
\usepackage{booktabs}
\usepackage{color}
\usepackage{siunitx}
\definecolor{AAA}{rgb}{1.0, 0.13, 0.32}

\biboptions{authoryear}

\begin{document}

\begin{frontmatter}

\title{D3Former: Jointly Learning Repeatable \textbf{D}ense \textbf{D}etectors and Feature-enhanced \textbf{D}escriptors via Saliency-guided Trans\textbf{former}}



\author[first]{Junjie Gao}
\ead{junjie.gao95.m@gmail.com}

\author[sec]{Pengfei Wang}
\ead{pf.wang.graphics@qq.com}

\author[first]{Qiujie Dong}
\ead{qiujie.jay.dong@gmail.com}

\author[first]{Qiong Zeng}
\ead{qiong.zn@sdu.edu.cn}

\cortext[cor]{Corresponding author}
\author[first]{Shiqing Xin*}
\ead{xinshiqing@sdu.edu.cn}

\author[last]{Caiming Zhang}
\ead{czhang@sdu.edu.cn}

\address[first]{School of Computer Science and Technology,
Shandong University, Qingdao, China}
\address[sec]{The University of Hong Kong, Pokfulam, Hong Kong}
\address[last]{School of Software,  Shandong University, Jinan, China}


\begin{abstract}
Establishing accurate and representative matches is a crucial step in addressing the point cloud registration problem. A commonly employed approach involves detecting keypoints with salient geometric features and subsequently mapping these keypoints from one frame of the point cloud to another. However, methods within this category are hampered by the repeatability of the sampled keypoints.
In this paper, we introduce a saliency-guided trans\textbf{former}, referred to as \textit{D3Former}, which entails the joint learning of repeatable \textbf{D}ense \textbf{D}etectors and feature-enhanced \textbf{D}escriptors. The model comprises a Feature Enhancement Descriptor Learning (FEDL) module and a Repetitive Keypoints Detector Learning (RKDL) module. The FEDL module utilizes a region attention mechanism to enhance feature distinctiveness, while the RKDL module focuses on detecting repeatable keypoints to enhance matching capabilities.
Extensive experimental results on challenging indoor and outdoor benchmarks demonstrate that our proposed method consistently outperforms state-of-the-art point cloud matching methods. Notably, tests on 3DLoMatch, even with a low overlap ratio, show that our method consistently outperforms recently published approaches such as RoReg and RoITr. For instance, with the number of extracted keypoints reduced to 250, the registration recall scores for RoReg, RoITr, and our method are 64.3\%, 73.6\%, and 76.5\%, respectively.

\end{abstract}


\end{frontmatter}


\section{Introduction}
Predicting an accurate rigid transformation between two unaligned partial point clouds, referred to as point cloud registration, constitutes a fundamental research problem with widespread applications in various fields such as 3D reconstruction~\cite{azinovic2022neural,deng2022depth,wang2023neural}, SLAM (Simultaneous Localization and Mapping)~\cite{Barros2022ACS,kong2023vmap,zhang2023go}, and autonomous driving\cite{Chitta2021NEATNA,Hu2022STP3EV}.
The common registration pipeline typically involves two steps: identifying representative point-to-point matches and estimating the final transformation. The challenge of point cloud registration persists, characterized by factors such as severe noise, weak textures, and low overlap rates, making it both intriguing and difficult.

In previous research, various dense matching methods~\cite{Yu2021CoFiNetRC, Qin2022GeometricTF, yu2023rotation, Choy2019FullyCG, Wang2021YouOH, Yu2022RIGARA} have been developed to extract a large number of candidate matches. However, these methods often face challenges in texture-less regions, where generating discriminative geometric features becomes difficult, leading to a significant number of mismatches.
While the coarse-to-fine paradigm~\cite{Yu2021CoFiNetRC, Qin2022GeometricTF, yu2023rotation} partially mitigates this issue, it is still prone to many mismatches that may collectively determine an incorrect transformation.

Recognizing the importance of points with high saliency in point cloud registration, detector-based matching methods~\cite{Yew20183DFeatNetWS, Li2019USIPUS, Bai2020D3FeatJL, Wu2020SKNetDL, huang2021predator} leverage detected keypoints to generate higher-quality matches. These methods can be categorized into two groups: detect-then-describe approaches~\cite{Yew20183DFeatNetWS, Li2019USIPUS, Wu2020SKNetDL} and detect-and-describe approaches~\cite{Bai2020D3FeatJL, huang2021predator}. The former detects representative keypoints and subsequently extracts features from local areas around each keypoint. On the other hand, the latter tightly integrates keypoint detection and descriptor learning. For instance, both D3Feat~\cite{Bai2020D3FeatJL} and Predator~\cite{huang2021predator} employ the kernel point convolutional neural network~\cite{Thomas2019KPConvFA} for joint detection and description, resulting in improved performance.
Despite their advancements, these methods treat the two point clouds independently, lacking a guarantee of the repeatability of sampled keypoints.

\begin{figure}[h]
  \centering
  \includegraphics[width=\linewidth]{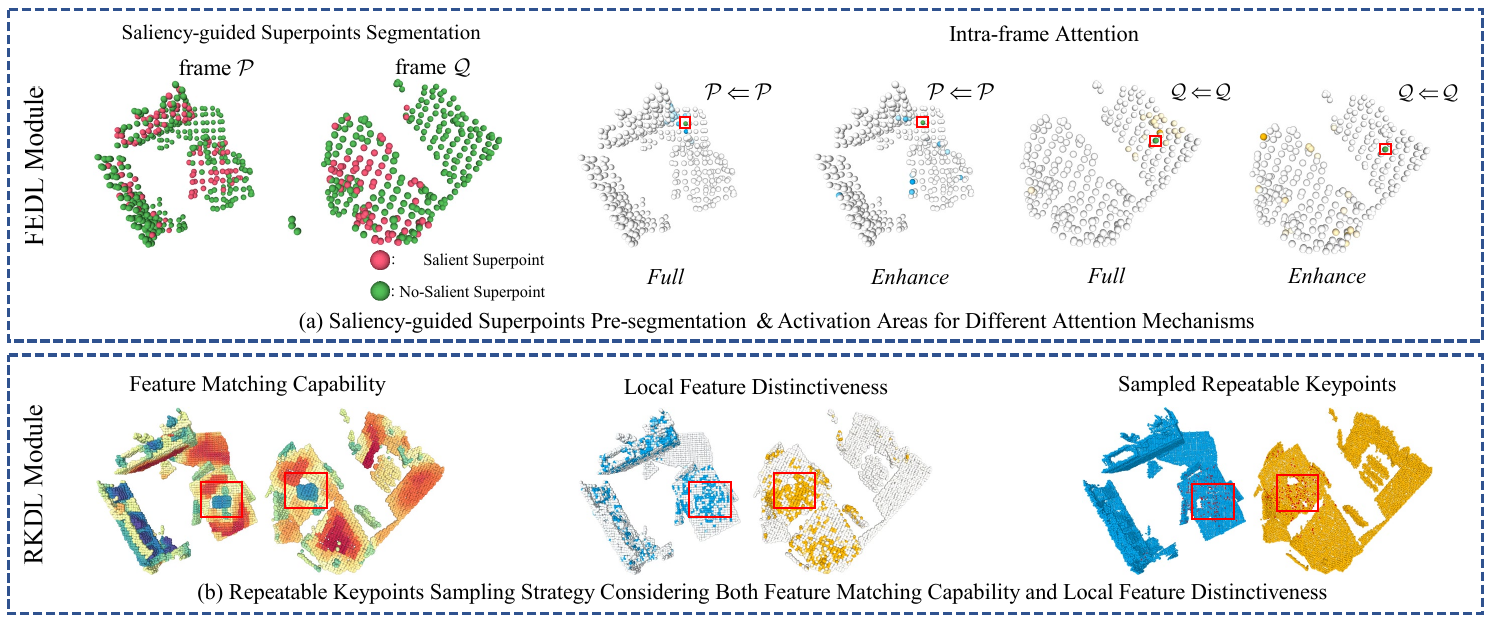}
  \vspace{-15pt}
  \caption{
 Two key modules in our network. (a) In the proposed FEDL module, we employ intra-frame enhanced attention to enhance the discriminative ability of superpoints, especially non-salient points.
(b) In the proposed RKDL module, we enhance both feature matching capability and local feature distinctiveness by sampling repeatable keypoints.
  }
  \label{fig.1}
\end{figure}

Building on the aforementioned discussion, it becomes evident that acquiring an accurate and representative set of matches is pivotal for effective point cloud registration. Enhancing the robustness of point cloud matching to real-world challenges involves careful consideration of the following two issues.
Motivated by insights gained from the literature review, we reevaluate two crucial problems. Firstly, \textbf{addressing the challenge of generating discriminative feature descriptors for texture-less regions is essential}. While some methods leverage kernel points~\cite{Thomas2019KPConvFA} or sparse voxel convolutions~\cite{Choy2019FullyCG} to extract point-wise features, the limited receptive field of these extracted features often hampers their discriminative ability in texture-less regions. Although certain approaches~\cite{Yew2022REGTREP,Qin2022GeometricTF,Yu2021CoFiNetRC,Li2021LepardLP} introduce full attention mechanisms to capture long-range dependencies, this may inadvertently lead to feature ambiguity.
Secondly, \textbf{enhancing the repeatability of sampled keypoints is another critical consideration}. A common challenging scenario arises when two point clouds overlap in a texture-less area. Relying solely on the saliency scores of keypoints for sampling can result in low repeatability. Therefore, it becomes imperative to devise a more intelligent keypoint sampling strategy to avoid blind sampling in such situations.

Motivated by the above observations, we propose a saliency-guided transformer named D3Former, consisting of a Feature Enhancement Descriptor Learning (FEDL) module and a Repetitive Keypoints Detection (RKDL) module.
The FEDL module aims to generate enhanced feature descriptors through a regional attention mechanism. It begins by extracting point-wise features based on the pre-trained model~\cite{Qin2022GeometricTF}, followed by computing salient scores based on feature differences. The regional attention mechanism facilitates a better understanding of the relationship between superpoints, enhancing the ability to perceive geometric structures and alleviating feature ambiguity.
The RKDL module focuses on detecting repeatable and well-matched keypoints. Initially, we identify candidate matches among superpoints with high matching capability through a dual softmax operation (typically for coarse levels)~\cite{cheng2021improving} or an optimal transport layer (typically for fine levels)~\cite{Sinkhorn1967ConcerningNM, cuturi2013sinkhorn}. These candidate matches undergo a scoring and filtering step to ensure keypoint repeatability and prevent sampling in texture-less patches. Finally, we obtain the final dense matches using a coarse-to-fine mechanism.

The main contributions can be summarized as follows:
\begin{itemize}
    \item We recognize two key factors for enhancing point cloud registration performance: generating discriminative feature descriptors for texture-less regions and improving the repeatability of sampled keypoints.
    \item We propose a saliency-guided transformer to jointly learn repeatable dense detectors and feature-enhanced descriptors. This encompasses a Feature Enhancement Descriptor Learning (FEDL) module to enhance feature discrimination and matching ability, and a Repetitive Keypoints Detection (RKDL) module to improve the repeatability of sampled keypoints.
    \item Extensive experimental results on challenging indoor and outdoor benchmarks reveal that our proposed method significantly outperforms both state-of-the-art detector-based and dense matching methods.
\end{itemize}

\section{Related Work}

In this section, we provide a brief overview of dense point cloud matching, detector-based point cloud matching, and the applications of transformers in vision-related tasks.

\textbf{Dense Point Cloud Matching.}
Dense point cloud matching methods typically consist of two stages: dense description and matching. Initially, these methods, considering all possible matches, rely on feature descriptors for efficient point-wise feature descriptions. Matches with high confidence are then obtained through the mutual nearest neighbor criterion. Traditional methods~\cite{Johnson1999UsingSI,Tombari2010UniqueSC,Tombari2010UniqueSO,Rusu2009FastPF,Guo20133DFF} construct descriptors using manually crafted features, limiting their adaptability to prior knowledge. Early learning-based feature descriptors faced challenges in achieving point-wise feature descriptors~\cite{Gojcic2018ThePM, Ao2020SpinNetLA, Zeng20163DMatchLL,deng2018ppfnet} due to computational complexity.

Subsequently, learning-based descriptors have been proposed, with FCGF~\cite{Choy2019FullyCG} standing out, utilizing sparse voxel convolution to achieve intensive feature computation with improved performance and lower computation cost. Additionally, KPConv~\cite{Thomas2019KPConvFA} proposed extracting point cloud features through a carefully designed 3D convolutional kernel and is often employed as the backbone of point cloud registration networks.

Recent learning-based dense matching methods~\cite{Li2021LepardLP, Yu2021CoFiNetRC, Qin2022GeometricTF, yu2023rotation} further improve matching quality by encoding and modeling long-range context information through an attention layer. They also apply an optimal transport layer~\cite{Sinkhorn1967ConcerningNM,cuturi2013sinkhorn} to identify more faithful matches. Despite notable performance improvements, the full attention mechanism may suffer from unnecessary correlation between salient regions and texture-less regions.
Based on our tests, there is an occurrence where many mismatches collectively report a wrong pose estimation. In our network, we encourage those non-salient superpoints in texture-less regions to correlate with relevant geometric structures, mitigating the risk of ambiguous features and contributing to more accurate pose estimations.

\textbf{Detector-based Point Cloud Matching.} 
Detector-based methods can be broadly categorized into two types: detect-then-describe~\cite{Yew20183DFeatNetWS,Li2019USIPUS,Wu2020SKNetDL} and detect-and-describe methods~\cite{Bai2020D3FeatJL,huang2021predator}. Detect-then-describe methods typically involve three stages: detection, description, and matching.
Recently, learning-based, data-driven approaches have been proposed for keypoint detection. One of the most representative works is USIP~\cite{Li2019USIPUS}, which adopts an unsupervised learning scheme that encourages keypoints to be covariant under arbitrary transformations. However, in these methods, keypoints and their feature descriptors are handled separately. A more integrated approach is pursued by detect-and-describe methods.

The earliest detect-and-describe work is D3Feat~\cite{Bai2020D3FeatJL}, which used KPConv~\cite{Thomas2019KPConvFA} as the backbone to extract point-wise features and guided keypoint detection through a self-supervision approach. Subsequently, Predator~\cite{huang2021predator} focused more on points located in overlapping regions, achieving considerable success on benchmarks with a low overlap rate. Despite this, detector-based matching methods rely on saliency scores as sampling criteria, which may be weak in ensuring the repeatability of sampled keypoints.

\textbf{Transformers in vision-related tasks.} Transformers~\cite{Vaswani2017AttentionIA} were initially prevalent in the natural language processing field, achieving notable success, such as with BERT~\cite{devlin2018bert}. Recently, Transformers have garnered increased attention in computer vision tasks, including object detection~\cite{chen2023diffusiondet}, image classification~\cite{dosovitskiy2020image}, and semantic segmentation~\cite{kirillov2023segment}.

DETR~\cite{carion2020end} employs a Transformer encoder-decoder architecture to replace handcrafted components, establishing the first fully end-to-end object detector with state-of-the-art performance. ViT~\cite{dosovitskiy2020image} explores the direct application of Transformers in image recognition with minimal modifications, dividing an image into non-overlapping patches as tokens and incorporating a specific positional encoding. These tokens are then input into standard Transformer layers to model global relations for image classification. PCT~\cite{guo2021pct}, based on Transformer's permutation invariance, directly applies the traditional Transformer architecture with 3D coordinate-based positional encoding and offset attention modules. This allows effective processing of unordered point clouds, achieving high performance in 3D shape classification, normal estimation, and segmentation tasks.

RoITr~\cite{yu2023rotation} introduces an attention mechanism embedding Point Pair Features (PPF)~\cite{Drost2010ModelGM} coordinates for describing pose-invariant geometry. It constructs a global Transformer with rotation-invariant cross-frame spatial awareness, significantly enhancing the rotation invariance of point cloud feature descriptions. Recently, Transformers have been introduced to the point cloud matching task with significant success, exemplified by CoFiNet~\cite{Yu2021CoFiNetRC} and Geotransformer~\cite{Qin2022GeometricTF}.

The superior performance of Transformer-based approaches can be attributed to the attention mechanism. Attention mechanisms are neural network layers with a long-range receptive field, adept at aggregating information from the entire input sequence. In this paper, we introduce attention mechanisms to the detector-based point cloud matching task, facilitating the learning of discriminative feature descriptors for texture-less regions.

\section{Method}
\begin{figure}[h]
  \centering
  \includegraphics[width=\linewidth]{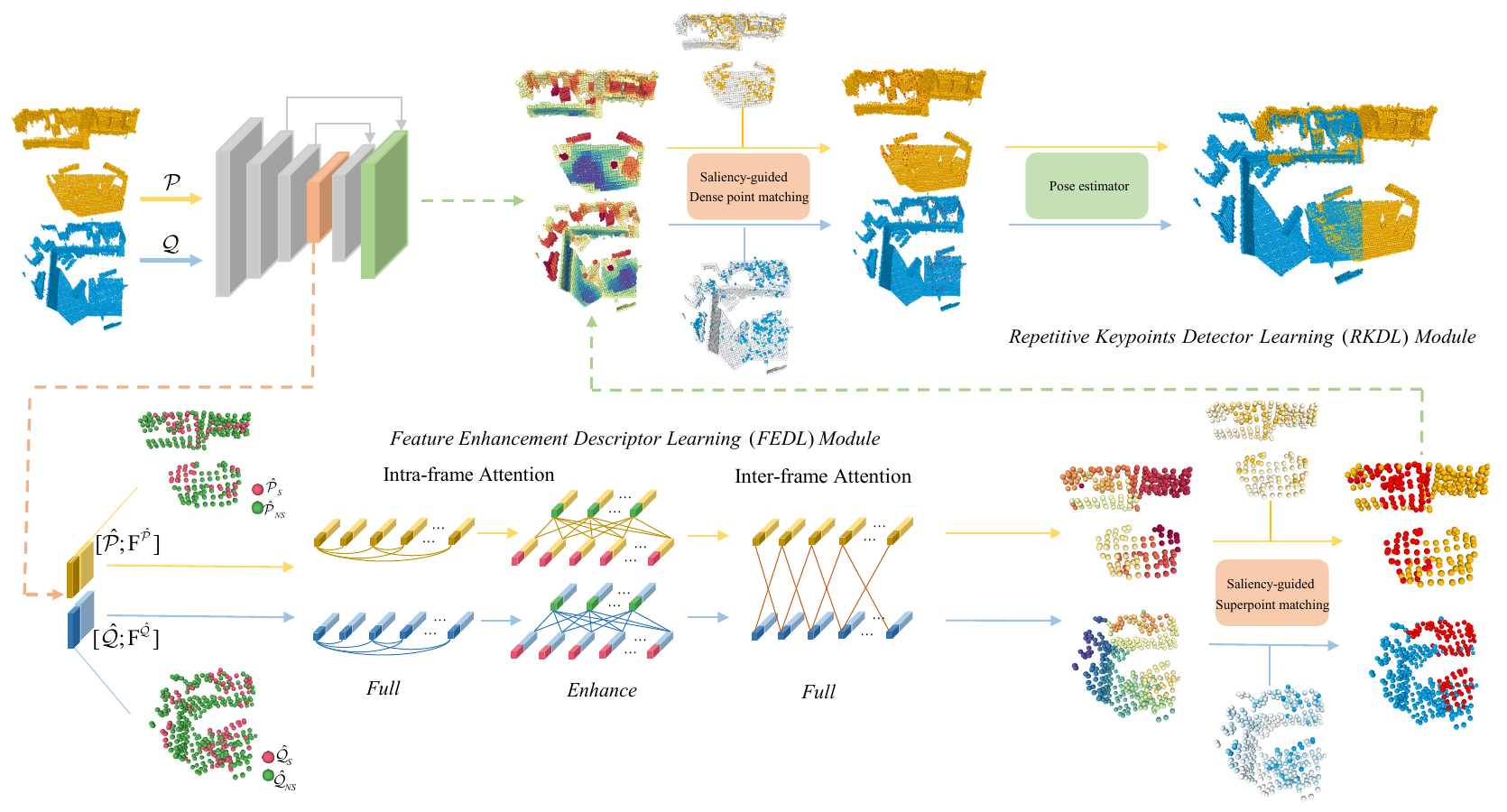}
  \vspace{-15pt}
  \caption{
  The architecture of our D3Former comprises two major components: a Feature Enhancement Descriptor Learning (FEDL) module and a Repetitive Keypoints Detector Learning (RKDL) module. The FEDL module encourages non-salient superpoints in texture-less regions to correlate with relevant geometric structures, thereby mitigating the risk of ambiguous features and contributing to more accurate pose estimations. The RKDL module aims to enhance the repeatability of sampled keypoints.
  }
  \label{fig.2}
\end{figure}

In this section, we present our proposed method by Jointly Learning Repeatable Dense Detectors and
Feature-Enhanced Descriptors via a Saliency-guided Transformer for point cloud matching.
The overall architecture is illustrated in Figure.~\ref{fig.2}.

\subsection{Overview}

Suppose we have a source point cloud $\mathcal{P}=\left\{\mathbf{p}_{i} \in \mathbb{R}^{3} \mid i=1, \ldots, N\right\}$ and a target point cloud $\mathcal{Q}=\left\{\mathbf{q}_{i} \in \mathbb{R}^{3}\mid i=1,\ldots, M\right\}$. The goal of rigid registration is to estimate the unknown rigid transformation $\mathbf{T}=\left\{\mathbf{R},\ \mathbf{t}\right\}$, where $\mathbf{R} \in \text{SO}(3)$ represents a rotation matrix, and $\mathbf{t} \in \mathbb{R}^{3}$ represents a translation vector.



To register the point clouds $\mathcal{P}$ and $\mathcal{Q}$, we employ a coarse-to-fine paradigm to establish matches. Our approach builds upon GeoTransformer~\cite{Qin2022GeometricTF}, utilizing KPConv~\cite{Thomas2019KPConvFA} to downsample the input point clouds and describe point-wise features simultaneously. The first and the last (coarsest) levels of downsampled points correspond to the dense points and the superpoints. We denote the superpoints as $\hat{\mathcal{P}}$ and $\hat{\mathcal{Q}}$, with the associated learned features as $\mathbf{F}^{\hat{\mathcal{P}}}$ and $\mathbf{F}^{\hat{\mathcal{Q}}}$.

We initially employ the pre-trained model to describe the features of superpoints. Subsequently, we score each superpoint based on its local feature differences. Finally, we categorize superpoints into salient superpoints $\hat{\mathcal{P}}_S$, $\hat{\mathcal{Q}}_S$ and non-salient superpoints $\hat{\mathcal{P}}_{NS}$, $\hat{\mathcal{Q}}_{NS}$ according to a specified proportion $\mathcal{N}$; refer to Sec.~\ref{sec.3.2}.


In the FEDL module, we employ a region attention mechanism to enhance feature distinctiveness within texture-less areas in the point cloud and improve the feature matching capability across salient regions in different point clouds. Specifically, besides utilizing geometric self-attention ($\hat{\mathcal{P}} \Leftarrow \hat{\mathcal{P}}$, $ \hat{\mathcal{Q}} \Leftarrow \hat{\mathcal{Q}}$) and full cross-attention ($\hat{\mathcal{P}} \Leftrightarrow \hat{\mathcal{Q}}$) to capture reliable long-range context and establish information exchange, we introduce a saliency-guided attention module. This module is designed to encode the correlation between non-salient superpoints and salient superpoints within a point cloud ($\hat{\mathcal{P}}_{NS} \Leftarrow \hat{\mathcal{P}}_S$, $ \hat{\mathcal{Q}}_{NS} \Leftarrow \hat{\mathcal{Q}}_S$); see Sec.~\ref{sec.3.3}.



In the RKDL module, we learn and obtain saliency scores ($\mathbf{S}^{\bar{\mathcal{P}}}$, $\mathbf{S}^{\bar{\mathcal{Q}}}$) for dense points in a self-supervised manner. The saliency score for each superpoint is defined as the average saliency score of dense points within its corresponding patch. Subsequently, we utilize the superpoint matching module of GeoTransformer~\cite{Qin2022GeometricTF} to extract superpoint matches, followed by a secondary filtering based on the obtained saliency scores. Within the dense point matching module, we apply a similar secondary filtering to the obtained dense matches, and the points on both sides of the resulting matches are used as keypoints for calculating the final transformation; see Sec.~\ref{sec.3.4}.

\subsection{Prior Saliency-Guided Segmentation} \label{sec.3.2}
Existing point cloud matching methods still face challenges, particularly in scenarios with low overlap and weak textures. The common approach is to use full attention~\cite{Vaswani2017AttentionIA,Li2021LepardLP,Yew2022REGTREP,Yu2021CoFiNetRC, Qin2022GeometricTF} to model long-range context. However, if the importance of different regions within the point cloud is not considered, such as establishing connections between points in weak texture regions and an excessive number of points in weak texture regions (planes), this can lead to a lack of connections with specific structural points (corners, edges), weakening the modeling capacity and causing feature blurring.
Therefore, we propose a prior saliency-guided segmentation method as a preprocessing step for point clouds. Our method is based on the observation that points in weak texture regions typically exhibit smaller local distinctiveness differences in feature descriptions, while conversely, points in strong texture regions have larger local distinctiveness differences.

Accordingly, we first extract features of superpoints using a pre-trained model and represent them as $\mathbf{F}^{\hat{\mathcal{P}}}$ and $\mathbf{F}^{\hat{\mathcal{Q}}}$. Inspired by detector-based methods, we follow D3Feat~\cite{Bai2020D3FeatJL} to calculate the saliency scores for each superpoint based on the local and channel distinctiveness of obtained features. The obtained scores are represented as $\mathbf{S}^{\hat{\mathcal{P}}}$ and $\mathbf{S}^{\hat{\mathcal{Q}}}$. Subsequently, we prioritize the salient regions by selecting the top-$k$ points from $\bar{\mathcal{P}}$ based on the descending order of their saliency scores $\mathbf{S}^{\bar{\mathcal{P}}}$. Here, $k$ is determined as $|\bar{\mathcal{P}}| \times \mathcal{N}$. The selected salient region points are denoted as $\hat{\mathcal{P}}_{S}$, while the non-salient region points are denoted as $\hat{\mathcal{P}}_{NS}$. Similarly, we apply the same procedure to compute $\hat{\mathcal{Q}}_{S}$ and $\hat{\mathcal{Q}}_{NS}$.

\subsection{Feature Enhancement Descriptor Learning} \label{sec.3.3}

To effectively mitigate feature ambiguity in texture-less regions and improve the matching capability of salient regions, we introduce a region attention mechanism in the proposed Feature Enhancement Descriptor Learning (FEDL) module.

For the intra-enhancement attention layer, keys and values are derived from the features of salient superpoints $\mathbf{F}^{\hat{\mathcal{P}}_{S}}$, while queries originate from the features of non-salient superpoints $\mathbf{F}^{\hat{\mathcal{P}}_{NS}}$.
Formally,
\begin{equation}
\mathbf{Q}= \mathbf{F}^{\hat{\mathcal{P}}_{NS}} \mathbf{W}^{Q}, \mathbf{K}=\mathbf{F}^{\hat{\mathcal{P}}_{S}}\mathbf{W}^{K} , \mathbf{V}=\mathbf{F}^{\hat{\mathcal{P}}_{S}}\mathbf{W}^{V} 
\end{equation}
where $\mathbf{W}^{Q} \in \mathbb{R}^{d \times d}, \mathbf{W}^{K} \in \mathbb{R}^{d \times d}, \mathbf{W}^{V} \in \mathbb{R}^{d \times d}$ are linear projections. Then, the non-salient superpoints features $\mathbf{F}^{\hat{\mathcal{P}}_{NS}}$ are updated in the following way,
\begin{equation}\label{eq.3}
\mathbf{A}^{\hat{\mathcal{P}}_{NS}}=\operatorname{Attention}(\mathbf{Q}, \mathbf{K}, \mathbf{V})=\operatorname{Softmax}\left(\mathbf{Q}\mathbf{K}^{\top}\right) \cdot \mathbf{V}
\end{equation}
\begin{equation}\label{eq.4}
\mathbf{F}^{\hat{\mathcal{P}}_{NS}}= \operatorname{LN}(\mathbf{F}^{\hat{\mathcal{P}}_{NS}} +\operatorname{MLP}(\mathbf{A}^{\hat{\mathcal{P}}_{NS}}))
\end{equation}
Motivated by ~\cite{Vaswani2017AttentionIA}, Eq.~\ref{eq.3} is implemented with the multihead attention. Here, LN represents layer normalization, and MLP denotes the multi-layer perception. The attention features for $\mathbf{F}^{\hat{\mathcal{Q}}_{NS}}$ are updated in the same way, while $\mathbf{F}^{\hat{\mathcal{P}}_{S}}$ and $\mathbf{F}^{\hat{\mathcal{Q}}_{S}}$ remain unchanged. In this manner, $\mathbf{F}^{\hat{\mathcal{P}}_{NS}}$ and $\mathbf{F}^{\hat{\mathcal{Q}}_{NS}}$ can effectively capture long-range context, establishing correlations between textureless areas and strong texture areas, thereby eliminating feature ambiguity.

\subsection{Repetitive Keypoints Detector Learning} \label{sec.3.4}

After obtaining context-enhanced features, our objective is to train a robust and reliable keypoints detector. This detector should not only detect repeatable keypoints in extreme scenarios but also identify keypoints when overlapping regions are entirely situated in texture-less areas. For this purpose, we consider not only the local feature distinctiveness as keypoints but also take into account the matching capability of features to ensure their repeatability.

In contrast to D3Feat~\cite{Bai2020D3FeatJL}, we did not conduct special training for the saliency score but directly obtained the saliency score of the obtained features. For the obtained dense point features $\mathbf{F}^{\bar{\mathcal{P}}}$ and $\mathbf{F}^{\bar{\mathcal{Q}}}$, we first calculate and obtain point-wise salient scores $\mathbf{S}^{\bar{\mathcal{P}}}$ and $\mathbf{S}^{\bar{\mathcal{Q}}}$ in the same way as D3Feat~\cite{Bai2020D3FeatJL}.

Subsequently, the saliency score for each superpoint is defined as the average saliency score of the dense points within its corresponding patch. The obtained superpoints' saliency scores are represented as $\mathbf{S}^{\hat{\mathcal{P}}}$ and $\mathbf{S}^{\hat{\mathcal{Q}}}$. Finally, the saliency scores obtained for these superpoints or dense points serve as guidance in establishing their respective matches. Below, we provide detailed explanations of saliency-guided matching for superpoints and dense points, along with some details on keypoint detection, followed by further discussions.

\textbf{Superpoints Matching.} Afterward, we normalize the enhanced superpoint features $\mathbf{F}^{\hat{\mathcal{P}}}$ and $\mathbf{F}^{\hat{\mathcal{Q}}}$ onto a unit hypersphere. We quantify pairwise similarity using a Gaussian correlation matrix $\mathcal{S}$ defined as follows:
\begin{equation}
\mathcal{S}(i, j) = \exp \left(-\left\|\mathbf{F}^{\hat{\mathcal{P}}}_i-\mathbf{F}^{\hat{\mathcal{Q}}}_j\right\|_2^2\right)
\end{equation}
Furthermore, we apply dual-normalization~\cite{cheng2021improving} on $\mathcal{S}$ to enhance global feature correlation. Superpoints associated with the top-$k$ entries are then selected as the superpoint matches set:
\begin{equation}
\hat{\mathcal{C}} = \left\{\left(\hat{\mathbf{p}}_i, \hat{\mathbf{q}}_j\right) \mid \hat{\mathbf{p}}_i \in \hat{\mathcal{P}}, \hat{\mathbf{q}}_j \in \hat{\mathcal{Q}}\right\}
\end{equation}
Based on the superpoints' saliency scores $\mathbf{S}^{\hat{\mathcal{P}}}$ and $\mathbf{S}^{\hat{\mathcal{Q}}}$, the saliency score for each match in the coarse match set is calculated as the sum of the saliency scores of the matched superpoints. The coarse matches set is refined by sorting the matches in descending order based on their saliency scores. The refined set, denoted as $\hat{\mathcal{C}}^*$, is obtained by selecting the top-$k$ matches. This final refined superpoint matches set captures the most salient and reliable matches between the two sets of features.

\textbf{Dense points Matching.} 
To establish dense points matches from $\hat{\mathcal{C}}^*$, dense points $\bar{\mathcal{P}}$ and $\bar{\mathcal{Q}}$ are initially assigned to superpoints. The point-to-node strategy~\cite{Li2018SONetSN} is leveraged for this purpose, where each point is assigned to its closest superpoint. Given a superpoint $\hat{\mathbf{p}}_i \in \hat{\mathcal{P}}$, the group of dense points assigned to it is denoted as $\mathcal{G}^{\hat{\mathcal{P}}}_i$ with $\mathcal{G}^{\hat{\mathcal{P}}}_i \subseteq \bar{\mathcal{P}}$. The group of features associated with $\mathcal{G}^{\hat{\mathcal{P}}}_i$ is further defined as $\mathcal{F}^{\hat{\mathcal{P}}}_i$ with $\mathcal{F}^{\hat{\mathcal{P}}}_i \subseteq \mathbf{F}^{\bar{\mathcal{P}}}$. For each superpoint match $\hat{\mathcal{C}}^*_l=\left(\hat{\mathbf{p}}_i, \hat{\mathbf{q}}_j\right)$, the similarity between the corresponding feature groups $\mathcal{F}^{\hat{\mathcal{P}}}_i$ and $\mathcal{F}^{\hat{\mathcal{Q}}}_j$ is calculated as:
\begin{equation}
\mathcal{M}_l=\mathcal{F}^{\hat{\mathcal{P}}}_{i}\left(\mathcal{F}^{\hat{\mathcal{Q}}}_{j}\right)^T / \sqrt{d}
\end{equation}
where $d$ is the feature dimension. 
We then follow~\cite{Wang2019SuperGLUEAS} to append a stack row and column to $\mathcal{M}_l$ filled with a learnable parameter. Subsequently, we iteratively run the Sinkhorn Algorithm~\cite{Sinkhorn1967ConcerningNM, cuturi2013sinkhorn} on it, yielding a normalized similarity matrix $\bar{\mathcal{M}}_l$. By removing the slack row and entry of $\bar{\mathcal{M}}_l$, we obtain the confidence matrix $\bar{\mathcal{C}}_l$. From this matrix, the mutual top-$k$ entries, i.e., entries with top-$k$ confidence on both the row and the column, are selected to formulate a dense point matches set $\mathcal{C}_l$. The final match set $\mathcal{C}$ is then collected as $\mathcal{C}=\cup_{l=1}^{|\mathcal{C}|} \mathcal{C}_l$.

Similarly, based on the dense points' saliency scores $\mathbf{S}^{\bar{\mathcal{P}}}$ and $\mathbf{S}^{\bar{\mathcal{Q}}}$, the dense match set $\mathcal{C}$ is further refined as $\mathcal{C}^*$. Within the dense match set $\mathcal{C}^*$, we denote the set of keypoints in the first dimension as $\mathcal{K}^{\mathcal{P}}=\left\{\mathbf{p}_i \mid \mathbf{p}_i \in \mathcal{C}^*\right\}$ representing key points on the $\mathcal{P}$ point cloud. Similarly, the set of keypoints in the second dimension is denoted as $\mathcal{K}^{\mathcal{Q}}=\left\{\mathbf{q}_j \mid \mathbf{q}_j \in \mathcal{C}^*\right\}$, representing keypoints on the $\mathcal{Q}$ point cloud.


\textbf{Discussion.}
Here, we discuss the differences between pre-trained models and our model in computing superpoint saliency scores. During the testing process, we utilize updated features and calculate the corresponding saliency scores instead of relying on pre-trained models. The primary reason for this choice is that the features obtained through our framework are more discriminative, enabling us to sample points with greater local distinctiveness. Meanwhile, during the training process, without any prior knowledge, we cannot preprocess the point cloud. Therefore, the pre-trained model is used exclusively for point cloud segmentation and guiding the training of our model.

\subsection{Loss Functions} \label{sec.3.5}
The final loss consists of the coarse-/fine-level loss and the overlap loss: $\mathcal{L} = \mathcal{L}_c + \mathcal{L}_f$. Similar to Geotransformer~\cite{Qin2022GeometricTF}, we use the overlap-aware circle loss $\mathcal{L}_c$~\cite{Qin2022GeometricTF} and negative log-likelihood loss $\mathcal{L}_f$~\cite{Wang2019SuperGLUEAS} for coarse and fine-level features, respectively. This allows features to be closer between superpoints/patches with higher overlap ratios in coarse-level matching, rather than strictly limiting one-to-one matching. At the fine level, stricter supervision also helps eliminate mismatches. For more details, please refer to~\cite{Qin2022GeometricTF}.

\section{Experiments}
In this section, we first present implementation details. Subsequently, we showcase experimental results and visualizations on two public benchmarks. Finally, a series of ablation studies are conducted to verify the effectiveness of each component.


\subsection{Implementation Details}
In this work, we implement the proposed model in PyTorch~\cite{Paszke2019PyTorchAI}. We use KPConv~\cite{Thomas2019KPConvFA} as the backbone to extract multi-level features. In the Saliency-Guided Segmentation module, we set $\mathcal{N}=0.35$ to divide superpoints into salient and non-salient categories. In the Feature Enhancement Descriptor Learning (FEDL) module, we repeat the FEDL module for $N_t=3$ times to achieve feature interaction and enhancement. In the Repetitive Keypoints Detector Learning (RKDL) module, we perform secondary filtering based on feature similarity (feat) and saliency score (saliency) at different phases to obtain corresponding matches. The hyperparameter $k$ determines the mutual top-$k$ selection in the dense matching module. Specific details are provided in Table~\ref{tab.1}. Other parameters remain unchanged according to~\cite{Qin2022GeometricTF}. For further details, please refer to the mentioned reference.

\begin{table}[h]
\centering
\caption{Detailed parameter settings of the RKDL module.}
\label{tab.1}
\scriptsize
\renewcommand{\arraystretch}{1.2}  
\setlength{\tabcolsep}{6pt}  
\begin{tabular}{cccccc} 
\toprule
           & \multicolumn{2}{c}{Superpoint matching} & \multicolumn{3}{c}{Dense point matching}  \\
\cmidrule(lr){2-3} \cmidrule(lr){4-6}
\# Samples & feat & saliency                          & $k$ & feat & saliency                  \\ 
\midrule
250        & 256  & 128                               & 1      & 500  & 250                       \\
500        & 256  & 128                               & 1      & 750  & 500                       \\
1000       & 256  & 192                               & 2      & 1500 & 1000                      \\
2500       & 256  & 256                               & 2      & 3500 & 2500                      \\
5000       & 256  & 256                               & 3      & All  & 5000                      \\
\bottomrule
\end{tabular}
\end{table}


\subsection{Datasets and Evaluation Metrics}
\textbf{3DMatch dataset.}
3DMatch comprises 62 indoor scenes, with 46 scenes used for training, 8 for validation, and 8 for testing. We preprocess the training data and evaluate our model on both the 3DMatch and 3DLoMatch benchmarks. The former features a 30\% overlap, while the latter exhibits low overlap in the range of 10\% to 30\%. To assess robustness to arbitrary rotations, we independently apply full-range rotations to the two frames of each point cloud pair.


For evaluation, we follow~\cite{Bai2020D3FeatJL,Yu2021CoFiNetRC,Qin2022GeometricTF} and employ four metrics: (1) {\em Inlier Ratio} (IR), which calculates the ratio of putative correspondences with a residual distance smaller than a threshold (0.1m) under the ground-truth transformation; (2) {\em Feature 
 Matching Recall} (FMR), which computes the fraction of point cloud pairs with an IR exceeding a threshold (5\%); (3) {\em Registration Recall} (RR), quantifying the fraction of point cloud pairs accurately registered (RMSE $<$ 0.2m); and (4) {\em Keypoint Repeatability} (KR), which calculates the ratio of sampled keypoints with a residual distance smaller than a threshold (0.1m) under the ground-truth transformation.


\textbf{KITTI dataset.}
The KITTI odometry dataset~\cite{Geiger2012AreWR} comprises 11 sequences of LiDAR-scanned outdoor driving scenarios. For training, we follow the setup of~\cite{Pais20193DRegNetAD,Bai2020D3FeatJL}, utilizing sequences 0$\sim$5. Sequences 6$\sim$7 are reserved for validation, and sequences 8-10 are designated for testing. Following the approach described in~\cite{huang2021predator}, we refine the ground-truth poses using ICP~\cite{Besl1992AMF} and restrict the evaluation to point cloud pairs within a maximum distance of 10 meters.

For evaluation metrics, we adhere to the criteria established by~\cite{Bai2020D3FeatJL,huang2021predator}, which include: (1) {\em Relative Rotation Error} (RRE): quantifying the geodesic distance between the estimated and ground-truth rotation matrices. (2) {\em Relative Translation Error} (RTE): calculating the Euclidean distance between the estimated and ground-truth translation vectors. (3) {\em Registration Recall} (RR): measuring the fraction of point cloud pairs for which both RRE and RTE fall below specific thresholds, typically set as RRE \textless $5^{\circ}$ and RTE \textless 2 meters.

\subsection{Comparison with State-of-the-art Methods}
\begin{figure*}[h]
  \centering
  \includegraphics[width=\linewidth]{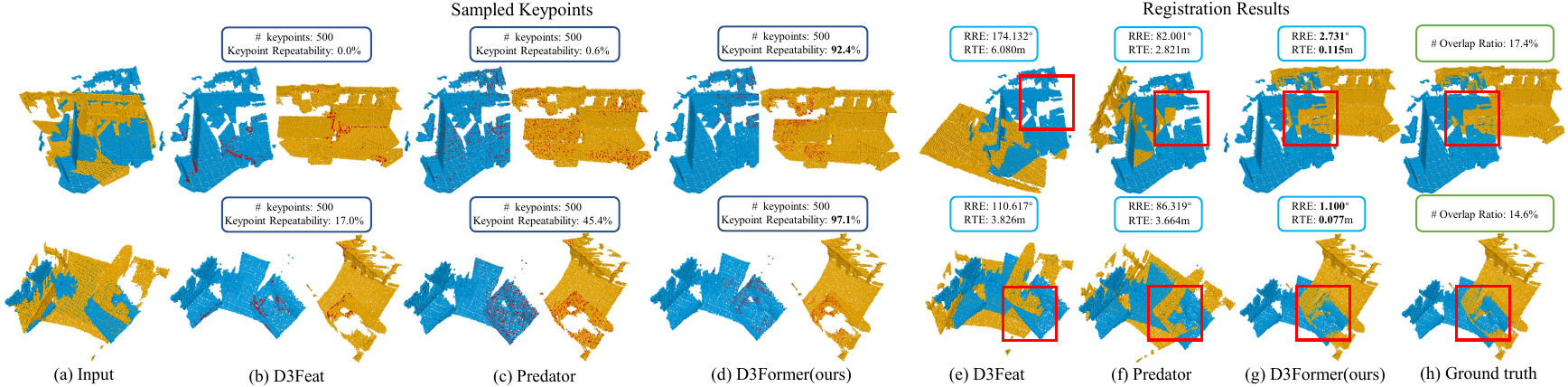}
  \vspace{-15pt}
   \caption{
   Qualitative results on 3DLoMatch. D3Feat and Predator are used as the baselines. Columns (b)–(d) display the sampled keypoints, while columns (e)–(g) illustrate the registration results. Red points represent keypoints.
  }
  \label{fig.3}
\end{figure*}

\begin{figure*}[h]
  \centering
  \includegraphics[width=\linewidth]{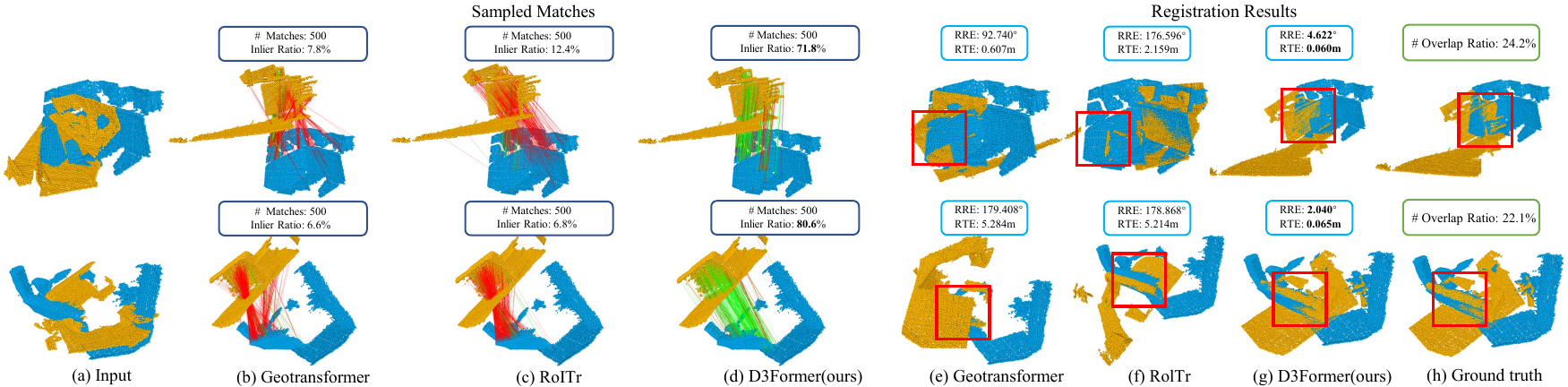}
  \vspace{-15pt}
  \caption{
  Qualitative results on 3DLoMatch. GeoTransformer and RolTr are used as the baselines. Columns (b)–(d) display the sampled matches, while columns (e)–(g) illustrate the registration results. Green lines represent inliers, and red lines represent outliers.
  }
  \label{fig.4}
\end{figure*}

\textbf{Results on 3DMatch dataset.} 
We compare our method with previous state-of-the-art point cloud matching methods~\cite{Gojcic2018ThePM, Choy2019FullyCG, Bai2020D3FeatJL,Ao2020SpinNetLA, huang2021predator,Wang2021YouOH, Yu2021CoFiNetRC, Qin2022GeometricTF, wang2023neural, yu2023rotation}, where D3Feat~\cite{Bai2020D3FeatJL} and Predator~\cite{huang2021predator} are detector-based sparse matching methods, and the rest are dense matching methods. We report the results with different numbers of matches in Table.~\ref{tab.2}, comparing with the best dense matching and detector-based methods. For FMR, when the sample number is less than 1k, our method achieves at least $0.5\% / 13.8\%$ improvement on 3DLoMatch, implying  under fewer samples, a higher likelihood of successful registration in scenarios with low overlap.  For IR, the improvements are even more prominent, consistently surpassing the baselines by $3.5\%\sim5.8\%$
on 3DMatch and $1.7\%\sim5.9\%$
on 3DLoMatch. This indicates that our obtained features have stronger discriminative capabilities, enabling the establishment of more accurate matches. It is noteworthy that our results are slightly lower than those of RoReg~\cite{Wang2023RoRegPP} and RolTr~\cite{yu2023rotation} when the number of sampled matches is 2.5k and 5k. This is attributed to the fact that these two methods employ point clouds at their original resolution, sampling more redundant correct matches in certain easily registerable scenarios, consequently achieving a higher inlier ratio (IR). In contrast, akin to GeoTransformer~\cite{Qin2022GeometricTF}, dense point matches are computed at 1/2 of the original resolution in our model. Under this configuration, compare with GeoTransformer~\cite{Qin2022GeometricTF}, we have increased the inlier ratio (IR) by at least 9.1\% and 6.4\% on 3DMatch and 3DLoMatch, respectively.

Registration recall performance, a crucial metric comprehensively reflecting the model's feature description quality, indicates the success rate of registration. Following~\cite{Bai2020D3FeatJL, huang2021predator}, we perform 50,000 RANSAC~\cite{Fischler1981RandomSC} iterations to estimate the transformation and find that our method consistently surpasses the baselines by $1.4\%\sim2.4\%$ on 3DMatch and $2.0\%\sim2.9\%$ on 3DLoMatch. Additionally, our method demonstrates stable performance under different numbers of samples, showcasing its robustness. 

\begin{table}[h]
\caption{Evaluation results on 3DMatch and 3DLoMatch with a varying number of matches.
}
\label{tab.2}
\vspace{2pt}
\setlength{\tabcolsep}{2.2pt}
\scriptsize
\centering
\begin{tabular}{l|ccccc|ccccc} 
\toprule
                         & \multicolumn{5}{c|}{3DMatch}                                                  & \multicolumn{5}{c}{3DLoMatch}                                                  \\
\# Samples               & 5000          & 2500          & 1000          & 500           & 250           & 5000          & 2500          & 1000          & 500           & 250            \\ 
\midrule
\multicolumn{11}{c}{\textit{Feature Matching Recall} (\%) $\uparrow$}                                                                                                                     \\ 
\midrule
PerfectMatch~\cite{Gojcic2018ThePM}            & 95.0          & 94.3          & 92.9          & 90.1          & 82.9          & 63.6          & 61.7          & 53.6          & 45.2          & 34.2           \\
FCGF~\cite{Choy2019FullyCG}                    & 97.4          & 97.3          & 97.0          & 96.7          & 96.6          & 76.6          & 75.4          & 74.2          & 71.7          & 67.3           \\
D3Feat~\cite{Bai2020D3FeatJL}                  & 95.6          & 95.4          & 94.5          & 94.1          & 93.1          & 67.3          & 66.7          & 67.0          & 66.7          & 66.5           \\
SpinNet~\cite{Ao2020SpinNetLA}                 & 97.6          & 97.2          & 96.8          & 95.5          & 94.3          & 75.3          & 74.9          & 72.5          & 70.0          & 63.6           \\
Predator~\cite{huang2021predator}                & 96.6          & 96.6          & 96.5          & 96.3          & 96.5          & 78.6          & 77.4          & 76.3          & 75.7          & 75.3           \\
YOHO~\cite{Wang2021YouOH}                    & \uline{98.2}  & 97.6          & 97.5          & 97.7          & 96.0          & 79.4          & 78.1          & 76.3          & 73.8          & 69.1           \\
CoFiNet~\cite{Yu2021CoFiNetRC}                 & 98.1          & \uline{98.3}  & 98.1          & \uline{98.2}  & \uline{98.3}  & 83.1          & 83.5          & 83.3          & 83.1          & 82.6           \\ 
GeoTrans~\cite{Qin2022GeometricTF}                & 97.9          & 97.9          & 97.9          & 97.9          & 97.6          & 88.3          & 88.6          & 88.8          & 88.6          & 88.3   \\
RoReg~\cite{Wang2023RoRegPP}                    & \uline{98.2}  & 97.9          & \uline{98.2}  & 97.8          & 97.2          & 82.1          & 82.1          & 81.7          & 81.6          & 80.2           \\
RoITr~\cite{yu2023rotation}                    & 98.0          & 98.0          & 97.9          & 98.0          & 97.9          & \uline{89.6} & \uline{89.6}  & \uline{89.5}  & \uline{89.4}  & \uline{89.3}   \\
D3Former (\textit{ours}) & \textbf{99.0} & \textbf{98.8} & \textbf{98.8} & \textbf{98.8} & \textbf{98.4} & \textbf{89.9} & \textbf{89.9} & \textbf{90.1} & \textbf{89.9} & \textbf{89.8}  \\ 
\midrule
\multicolumn{11}{c}{\textit{Inlier Ratio} (\%) $\uparrow$}                                                                                                                                \\ 
\midrule
PerfectMatch~\cite{Gojcic2018ThePM}               & 36.0          & 32.5          & 26.4          & 21.5          & 16.4          & 11.4          & 10.1          & 8.0           & 6.4           & 4.8            \\
FCGF~\cite{Choy2019FullyCG}                       & 56.8          & 54.1          & 48.7          & 42.5          & 34.1          & 21.4          & 20.0          & 17.2          & 14.8          & 11.6           \\
D3Feat~\cite{Bai2020D3FeatJL}                    & 39.0          & 38.8          & 40.4          & 41.5          & 41.8          & 13.2          & 13.1          & 14.0          & 14.6          & 15.0           \\
SpinNet~\cite{Ao2020SpinNetLA}                 & 47.5          & 44.7          & 39.4          & 33.9          & 27.6          & 20.5          & 19.0          & 16.3          & 13.8          & 11.1           \\
Predator~\cite{huang2021predator}                & 58.0          & 58.4          & 57.1          & 54.1          & 49.3          & 26.7          & 28.1          & 28.3          & 27.5          & 25.8           \\
YOHO~\cite{Wang2021YouOH}                    & 64.4          & 60.7          & 55.7          & 46.4          & 41.2          & 25.9          & 23.3          & 22.6          & 18.2          & 15.0           \\
CoFiNet~\cite{Yu2021CoFiNetRC}                 & 49.8          & 51.2          & 51.9          & 52.2          & 52.2          & 24.4          & 25.9          & 26.7          & 26.8          & 26.9           \\
GeoTrans~\cite{Qin2022GeometricTF}                 & 71.9          & 75.2          & 76.0          & 82.2          & \uline{85.1}  & 43.5          & 45.3          & 46.2          & 52.9          & \uline{57.7}   \\
RoReg~\cite{Wang2023RoRegPP}                      & \uline{81.6}          & 80.2          & 75.1          & 74.1          & 75.2          & 39.6          & 39.6          & 34.0          & 31.9          & 34.5           \\
RoITr~\cite{yu2023rotation}                     & \textbf{82.6} & \uline{82.8}  & \uline{83.0}  & \uline{83.0}  & 83.0          & \textbf{54.3} & \textbf{54.6} & \uline{55.1}  & \uline{55.2}  & 55.3  \\
D3Former (\textit{ours}) & 81.0  & \textbf{85.6} & \textbf{86.5} & \textbf{88.0} & \textbf{88.8} & \uline{50.4}  & \uline{51.4} & \textbf{56.8} & \textbf{59.5} & \textbf{61.2}  \\ 
\midrule
\multicolumn{11}{c}{\textit{Registration Recall} (\%) $\uparrow$}                                                                                                                         \\ 
\midrule
PerfectMatch~\cite{Gojcic2018ThePM}             & 78.4          & 76.2          & 71.4          & 67.6          & 50.8          & 33.0          & 29.0          & 23.3          & 17.0          & 11.0           \\
FCGF~\cite{Choy2019FullyCG}                      & 85.1          & 84.7          & 83.3          & 81.6          & 71.4          & 40.1          & 41.7          & 38.2          & 35.4          & 26.8           \\
D3Feat~\cite{Bai2020D3FeatJL}                    & 81.6          & 84.5          & 83.4          & 82.4          & 77.9          & 37.2          & 42.7          & 46.9          & 43.8          & 39.1           \\
SpinNet~\cite{Ao2020SpinNetLA}                 & 88.6          & 86.6          & 85.5          & 83.5          & 70.2          & 59.8          & 54.9          & 48.3          & 39.8          & 26.8           \\
Predator~\cite{huang2021predator}                & 89.0          & 89.9          & 90.6          & 88.5          & 86.6          & 59.8          & 61.2          & 62.4          & 60.8          & 58.1           \\
YOHO~\cite{Wang2021YouOH}                   & 90.8          & 90.3          & 89.1          & 88.6          & 84.5          & 65.2          & 65.5          & 63.2          & 56.5          & 48.0           \\
CoFiNet~\cite{Yu2021CoFiNetRC}                  & 89.3          & 88.9          & 88.4          & 87.4          & 87.0          & 67.5          & 66.2          & 64.2          & 63.1          & 61.0           \\
GeoTransr~\cite{Qin2022GeometricTF}                 & 92.0          & 91.8          & 91.8          & 91.4          & \uline{91.2}  & \uline{75.0}  & \uline{74.8}  & 74.2          & 74.1          & 73.5   \\
RoReg~\cite{Wang2023RoRegPP}                      & \uline{92.9}  & \uline{93.2}  & \uline{92.7}  & \uline{93.3}  & \uline{91.2}  & 70.3          & 71.2          & 69.5          & 67.9          & 64.3           \\
RoITr~\cite{yu2023rotation}                       & 91.9          & 91.7          & 91.8          & 91.4          & 91.0          & 74.7          & \uline{74.8}  & \uline{74.8}  & \uline{74.2}  & \uline{73.6}   \\
D3Former (\textit{ours}) & \textbf{94.5} & \textbf{94.6} & \textbf{94.2} & \textbf{94.7} & \textbf{93.6} & \textbf{77.0} & \textbf{77.0} & \textbf{77.2} & \textbf{76.7} & \textbf{76.5}  \\
\bottomrule
\end{tabular}
\end{table}


We also compare the registration recall of the RANSAC-free estimator in Table.~\ref{tab.3}. Due to our higher FMR and IR, we achieved better performance by extracting a set of matches with high matching capability within specific geometric structures. We start with weighted SVD over 250 matches
in solving for pose transformation, which imposes higher requirements on the model's feature matching recall and inlier ratio. Our method achieves registration recall rates of 89.4\% and 63.5\% on 3DMatch and 3DLoMatch, respectively, close to CoFiNet~\cite{Yu2021CoFiNetRC} with RANSAC~\cite{Fischler1981RandomSC}. In addition, we also use LGR~\cite{Qin2022GeometricTF} as the pose estimator, our method achieved registration recall rates of 92\% and 80\% on 3DMatch and 3DLoMatch, respectively. This performance is comparable to the RANSAC-based version but with faster speed, demonstrating that our method is applicable for accelerating the registration process.

Finally, qualitative comparison results with detector-based and dense-matching methods are presented in Figure.~\ref{fig.3} and Figure.~\ref{fig.4}, respectively. Compared to detector-based methods, our proposed keypoint sampling strategy, considering both saliency and feature matching capability, results in keypoints consistently located in overlapping regions with higher repeatability. Even in scenarios where the overlapping areas of point clouds are entirely within texture-missing regions, our method can still sample repeatable keypoints, achieving more robust matching. Compared to dense matching methods, the proposed saliency-guided attention mechanism significantly enhances the feature discriminability of points in weakly textured areas and the feature matching capability of points in strongly textured areas. Even in extreme scenarios, we can establish accurate matches, fully demonstrating the effectiveness of our method.

\begin{table}[h]
\caption{Registration results w/o RANSAC on 3DMatch (3DM) and 3DLoMatch (3DLM).}
\label{tab.3}
\vspace{2pt}
\setlength{\tabcolsep}{5pt}
\scriptsize
\centering
\begin{tabular}{l|l|c|cc} 
\toprule
\multirow{2}{*}{Model} & \multicolumn{1}{c|}{\multirow{2}{*}{Estimator}} & \multirow{2}{*}{\#Samples} & \multicolumn{2}{c}{RR(\%)}     \\
                       & \multicolumn{1}{c|}{}                           &                            & \multicolumn{2}{c}{3DM 3DLM}   \\ 
\midrule
SpinNet~\cite{Ao2020SpinNetLA}               & RANSAC-50k                                      & 5000                       & 88.6          & 59.8           \\
Predator~\cite{huang2021predator}              & RANSAC-50k                                      & 5000                       & 89.0          & 59.8           \\
CoFiNet~\cite{Yu2021CoFiNetRC}               & RANSAC-50k                                      & 5000                       & 89.3          & 67.5           \\
GeoTrans~\cite{Qin2022GeometricTF}              & RANSAC-50k                                      & 5000                       & \uline{92.0}  & \uline{75.0}   \\
RoITr~\cite{yu2023rotation}                  & RANSAC-50k                                      & 5000                       & 91.9          & 74.7           \\
D3Former(ours)         & RANSAC-50k                                      & 5000                       & \textbf{94.2} & \textbf{77.2}  \\ 
\midrule
SpinNet~\cite{Ao2020SpinNetLA}               & weighted SVD                                    & 250                        & 34.0          & 2.5            \\
Predator~\cite{huang2021predator}               & weighted SVD                                    & 250                        & 50.0          & 6.4            \\
CoFiNet~\cite{Yu2021CoFiNetRC}               & weighted SVD                                    & 250                        & 64.6          & 21.6           \\
GeoTrans~\cite{Qin2022GeometricTF}              & weighted SVD                                    & 250                        & \uline{86.5}  & 59.9           \\
RolTr~\cite{yu2023rotation}                   & weighted SVD                                    & 250                        & \uline{86.5}  & \uline{60.0}   \\
D3Former(ours)         & weighted SVD                                    & 250                        & \textbf{89.4} & \textbf{63.5}  \\ 
\midrule
CoFiNet~\cite{Yu2021CoFiNetRC}               & \multicolumn{1}{c|}{LGR}                        & all                        & 85.5          & 63.2           \\
GeoTrans~\cite{Qin2022GeometricTF}              & \multicolumn{1}{c|}{LGR}                        & all                        & \uline{91.5}  & 74.0           \\
RolTr~\cite{yu2023rotation}                  & \multicolumn{1}{c|}{LGR}                        & all                        & 91.0          & \uline{73.6}   \\
D3Former(ours)         & \multicolumn{1}{c|}{LGR}                        & all                        & \textbf{94.3} & \textbf{76.2}  \\
\bottomrule
\end{tabular}
\vspace{-5pt}
\end{table}

The repeatability of keypoints is a crucial metric for evaluating keypoint detectors. Therefore, we compared our method with detector-based approaches~\cite{Bai2020D3FeatJL,huang2021predator} in Table.~\ref{tab.4}. Following with ~\cite{Bai2020D3FeatJL}, for each detector, we generated 4, 8, 16, 32, 64, 128, 256, 512 keypoints and calculated the relative repeatability for each method. Due to our method considering both saliency and feature similarity, it significantly improves the KR of sampled keypoints. Specifically, it consistently outperforms the baseline by $16.8\%\sim71.6\%$ on 3DMatch and $22.6\%\sim60.1\%$ on 3DLoMatch.

\begin{table}[h]
\centering
\caption{Keypoints repeatability on 3DMatch and 3DLoMatch with a varying sample number.}
\label{tab.4}
\setlength{\tabcolsep}{2pt}
\scriptsize
\centering
\begin{tabular}{lcccccccc} 
\toprule
& \multicolumn{8}{c}{KR(\%) 3DMatch/3DLoMatch} \\
\cmidrule(lr){2-9}
\# Samples & 4 & 8 & 16 & 32 & 64 & 128 & 256 & 512 \\
\midrule
D3Feat~\cite{Bai2020D3FeatJL} & \uline{20.6}/\uline{6.6} &\uline{24.3}/\uline{8.5} & \uline{29.1}/\uline{11.1} & 34.3/13.6 & 39.4/16.9 & 44.2/20.5 & 49.0/24.3 & 52.7/27.7 \\
Predator~\cite{huang2021predator} & 3.3/2.5  & 6.0/4.8  & 11.3/10.0 & \uline{20.5}/\uline{17.3} & \uline{34.3}/\uline{28.2} & \uline{51.9}/\uline{40.3} & \uline{68.3}/\uline{49.8} & \uline{79.3}/\uline{55.6} \\
D3Former (ours) &\textbf{92.2}/\textbf{66.7} &\textbf{92.7}/\textbf{68.6} &\textbf{93.1}/\textbf{70.5} &\textbf{93.7}/\textbf{72.5} &\textbf{94.6}/\textbf{74.7} &\textbf{95.3}/\textbf{76.6} &\textbf{95.7}/\textbf{77.7} &\textbf{96.1}/\textbf{78.2}  \\
\bottomrule
\end{tabular}
\vspace{-15pt}
\end{table}

\begin{table}[h]
\caption{Registration results on KITTI odometry.}
\label{tab.5}
\scriptsize
\setlength{\tabcolsep}{4pt}
\centering
\begin{tabular}{l|ccc}
\toprule
Model & RTE(cm) & RRE($^{\circ}$) & RR(\%) \\
\midrule
3DFeat-Net~\cite{Yew20183DFeatNetWS} & 25.9 & \textbf{0.25} & 96.0 \\
FCGF~\cite{Choy2019FullyCG} & 9.5 & 0.30 & 96.6 \\
D3Feat~\cite{Bai2020D3FeatJL} & 7.2 & 0.30 & \textbf{99.8} \\
SpinNet~\cite{Ao2020SpinNetLA} & 9.9 & 0.47 & 99.1 \\
Predator~\cite{huang2021predator} & 6.8 & 0.27 & \textbf{99.8} \\
CoFiNet~\cite{Yu2021CoFiNetRC} & 8.2 & 0.41 & \textbf{99.8} \\
GeoTrans~\cite{Qin2022GeometricTF} & \underline{7.4} & 0.27 & \textbf{99.8} \\
D3Former (\emph{ours}, RANSAC-\emph{50k}) & \textbf{6.8} & \textbf{0.25} & \textbf{99.8} \\
\midrule
FMR~\cite{Huang2020FeatureMetricRA} & $\sim$66 & 1.49 & 90.6 \\
DGR~\cite{Choy2020DeepGR}& $\sim$32 & 0.37 & 98.7 \\
HRegNet~\cite{Lu2021HRegNetAH}& $\sim$12 & 0.29 & 99.7 \\
GeoTrans (LGR)~\cite{Qin2022GeometricTF} & \underline{6.8} & \underline{0.24} & \textbf{99.8} \\
D3Former (\emph{ours}, LGR) & \textbf{6.2} & \textbf{0.22} & \textbf{99.8} \\
\bottomrule
\end{tabular}
\vspace{-5pt}
\label{table:kitti}
\end{table}

\textbf{Results on KITTI dataset.} 
As shown in Table.~\ref{tab.5}, we compare our method with previous state-of-the-art approaches~\cite{Yew20183DFeatNetWS, Choy2019FullyCG, Bai2020D3FeatJL, Ao2020SpinNetLA, huang2021predator, Yu2021CoFiNetRC, Qin2022GeometricTF} to validate the effectiveness of our D3Former for outdoor pose estimation. We include comparisons with both detector-based methods and dense matching methods. Notably, our approach achieves lower relative rotation and translation errors while maintaining comparable registration recall rates. This strongly demonstrates the generalization capability of our method in outdoor scenes.

\begin{figure*}[h]
  \centering
  \includegraphics[width=\linewidth]{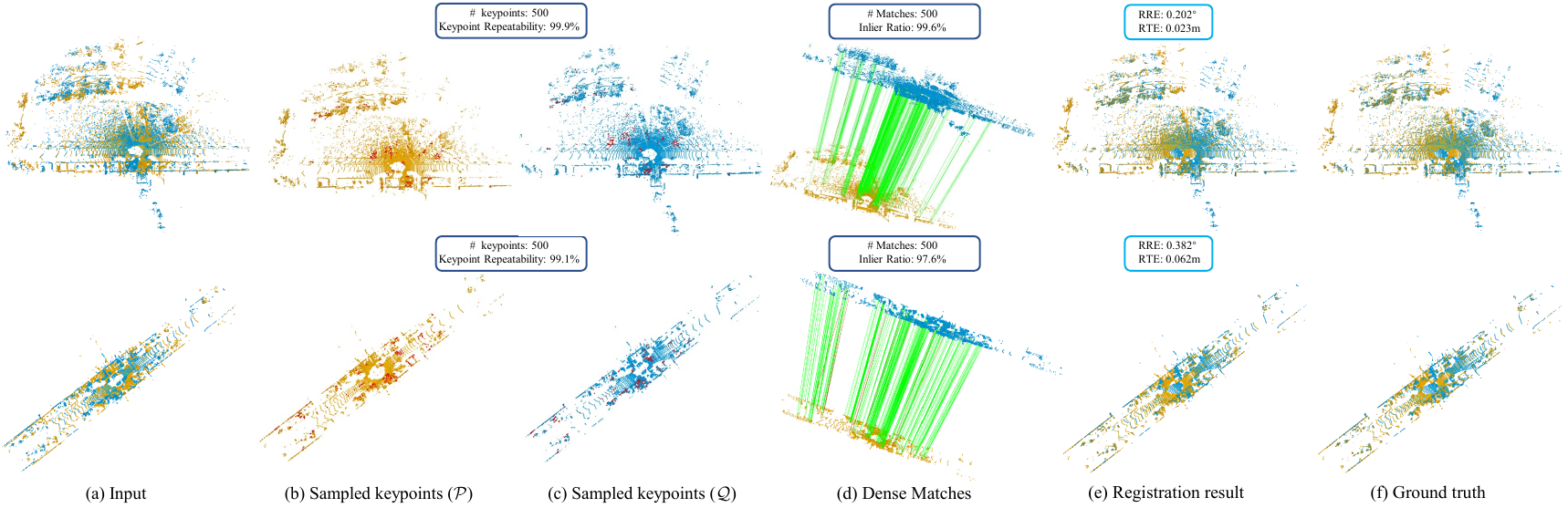}
  \vspace{-15pt}
  \caption{
  Qualitative results on KITTI odometry. Columns (b) and (c) display the sampled keypoints. Column (d) show the matches while (e) illustrate registration results. Green lines represent inliers, and red lines represent outliers.
  }
  \label{fig.5}
\end{figure*}

Furthermore, we compare with RANSAC-free methods~\cite{Huang2020FeatureMetricRA, Choy2020DeepGR, Lu2021HRegNetAH, Qin2022GeometricTF}, utilizing LGR as the pose estimator. It is evident that, benefiting from the precise matches obtained, similar to~\cite{Qin2022GeometricTF}, our method facilitates accelerated outdoor scene registration. Additionally, qualitative experimental results for outdoor scenes are presented in Figure.~\ref{fig.5}. Despite the sparse nature of outdoor point clouds, our method can sample keypoints in specific geometric structure regions and establish accurate matches, achieving robust registration with low RRE and RTE.


\subsection{Ablation Studies}
To analyze the effects of each component in D3Former, we conduct detailed ablation studies on the 3DMatch and 3DLoMatch datasets. To evaluate superpoint (patch) matching, we follow~\cite{Qin2022GeometricTF} and introduce another metric Patch Inlier Ratio (PIR), which is the fraction of patch matches with actual overlap. The metrics FMR and IR are reported with all dense point matches, and LGR is employed as the pose estimator for registration.

In Table.~\ref{tab.6}, the model [A] is the same structure as the Geotransformer~\cite{Qin2022GeometricTF}. Subsequently, for Model [B], only the feature enhancement descriptor learning (FEDL) module is added to eliminate the feature ambiguity of non-significant superpoints. For Model [C], only the repetitive keypoints detector learning (RKDL) module is added to find a set of accurate and representative matches. The model [D] is the complete model of our D3Former.

\begin{table}
\centering
\caption{Effectiveness of each component on 3DMatch and 3DLoMatch.}
\label{tab.6}
\scriptsize
\setlength{\tabcolsep}{5pt}
\centering
\begin{tabular}{ccc|S[table-format=2.1]S[table-format=2.1]S[table-format=2.1]S[table-format=2.1]|S[table-format=2.1]S[table-format=2.1]S[table-format=2.1]S[table-format=2.1]} 
\toprule
      &      &      & \multicolumn{4}{c|}{3DMatch}                                                                         & \multicolumn{4}{c}{3DLoMatch}                                                                         \\
\cmidrule(lr){4-7} \cmidrule(lr){8-11}
Model & FEDL & RKDL & {PIR} & {FMR} & {IR} & {RR} & {PIR} & {FMR} & {IR} & {RR}  \\ 
\midrule
{[}A] &  $\times$  & $\times$      &  86.1     &       97.7     &        70.3                &      91.5        &   54.9      &     88.1       &        43.3                 &    74.0                     \\
{[}B] &  $\checkmark$  & $\times$      &     \underline{87.9}   & \underline{98.8}         &  \underline{76.1}     &\underline{94.0}   & \underline{57.6}  & \textbf{89.2}        & \underline{50.2}   & \underline{75.8}  \\ 
{[}C] &  $\times$   & $\checkmark$    &  86.6      &   98.0         &  73.8                      &   91.8           & 55.4         &   88.3    &  45.8                       &    74.8                     \\
{[}D] &  $\checkmark$  & $\checkmark$     &    \textbf{88.4}   & \textbf{99.0}           &  \textbf{79.2}   & \textbf{94.3}           &  \textbf{58.0}  & \textbf{89.2}   
  &\textbf{53.1} &  \textbf{76.2}         \\
\bottomrule
\end{tabular}
\end{table}

\textbf{Effects of the feature enhancement descriptor learning (FEDL) module.}
As shown in Table.~\ref{tab.6}, with the proposed FEDL module, the performance on the 3DMatch and 3DLoMatch is improved
notably. In specific, compared to the model [A], the performance of model [B] is improved by 1.8\%/2.7\% in PIR, 1.1\%/1.1\% in FMR, 5.8\%/6.9\% in IR and 2.5\%/1.8\% in RR. And the model [D] also outperforms model [C], particularly evident in terms of PIR and IR. It can be observed that by model long-range context between non-salient superpoints and salient superpoints, our model effectively eliminates the ambiguity in the features of non-salient superpoints, thereby enhancing their feature distinctiveness and matching capability. Consequently, this leads to the establishment of more accurate matches, resulting in a higher RR.

\begin{figure*}[h]
  \centering
  \includegraphics[width=\linewidth]{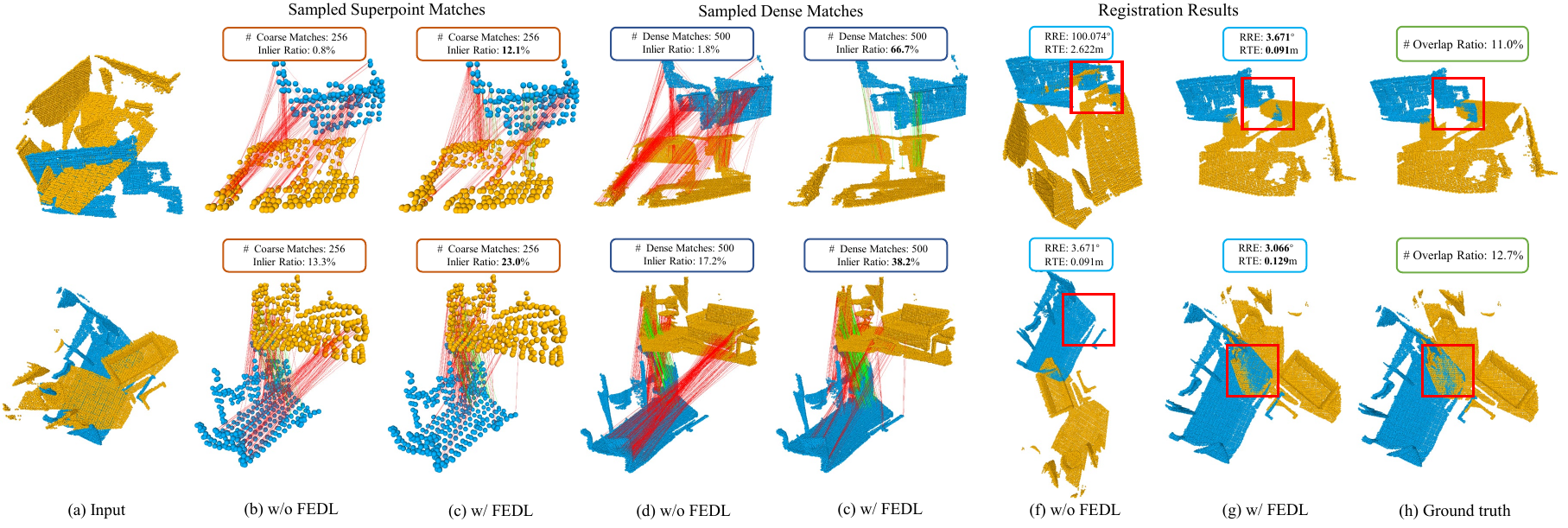}
  \vspace{-15pt}
  \caption{
    Qualitative Comparison of Results with (w/) and without (w/o) the FEDL Module. Columns (b) and (c) display the sampled superpoint matches. Columns (d) and (e) display the sampled dense point matches, while columns (f) and (g) illustrate the registration results. Green lines represent inliers, and red lines represent outliers.
  }
  \label{fig.6}
\end{figure*}

Additionally, we qualitatively compared models trained without (w/o) and with (w/) the FEDL module, confirming the effectiveness of the designed intra-frame feature enhancement attention mechanism. As shown in Figure.~\ref{fig.6}, we observed that, due to the enhanced feature distinctiveness and matching capability of non-salient superpoints, some superpoints in texture-less areas are accurately matched and propagated to dense point matching. This resulted in a higher inlier ratio and more accurate registration results.

\textbf{Impacts about the pre-segmention proportion $\mathcal{N}$ in the FEDL module.} Here, we study the performance with different pre-segmention proportion ($\mathcal{N}$) in the FEDL module. We uniformly sample $\mathcal{N}$ values within a reasonable range (0.1$\sim$0.5) and evaluate the performance on 3DMatch and 3DLoMatch. As shown in Table.~\ref{tab.7}, we observe that the model achieves optimal performance when $\mathcal{N}$ is set to 0.35. This may be because modeling non-salient points along with too small a proportion of salient points is insufficient to eliminate feature ambiguity, while interacting with too large a proportion of salient points typically introduces too much irrelevant information, adversely affecting model training.

\textbf{Effects of the repetitive keypoints detector learning (RKDL) module.}
As shown in Table.~\ref{tab.6}, when adding our proposed RKDL module, the performance on the 3DMatch and 3DLoMatch can achieve a kind of improvement. Specifically, compare to the model [A], the performance of model [C] is gained by 0.5\%/0.5\% in PIR, 0.3\%/0.2\% in FMR, 3.5\%/2.5\% in IR and 0.3\%/0.8\% in RR. Besides, the model [D] also performs better than the model [B]. The primary reason is that our proposed RKDL module can identify a representative set of matches, thereby avoiding clustering together and preventing falling into local optima. This is crucial for robust point cloud registration. Similar discussions on this aspect can also be found in Predator~\cite{huang2021predator}.

\begin{table}
\centering
\caption{Impacts of the pre-segmentation Proportion $\mathcal{N}$ on 3DMatch and 3DLoMatch.}
\label{tab.7}
\scriptsize
\setlength{\tabcolsep}{6pt}
\centering
\begin{tabular}{c|cccc|cccc} 
\toprule
\multirow{3}{*}{Model} & \multicolumn{4}{c|}{3DMatch} & \multicolumn{4}{c}{3DLoMatch}  \\
\cmidrule(lr){2-5} \cmidrule(lr){6-9}
& PIR & FMR & IR & RR & PIR & FMR & IR & RR \\ 
\midrule
$\mathcal{N}$=0.10 & 85.7 & 98.0 & 72.7 & 92.8 & 54.0 & 88.4 & 46.2 & 74.5 \\
$\mathcal{N}$=0.15 & 86.1 & 98.2 & 74.2 & 93.0 & 54.3 & 88.6 & 46.7 & 74.9 \\
$\mathcal{N}$=0.20 & 86.5 & 98.4 & 75.5 & 93.4 & 55.3 & 88.9 & 47.8 & 75.2 \\
$\mathcal{N}$=0.25 & 87.2 & 98.6 & 75.7 & 93.7 & 56.6 & 88.9 & 47.7 & 75.2 \\
$\mathcal{N}$=0.30 & 87.4 & \textbf{98.8} & 75.9 & 93.7 & \uline{57.2} & \textbf{89.2} & \uline{48.3} & \uline{75.6} \\
$\mathcal{N}$=0.35 & \uline{87.9} & \textbf{98.8} & \uline{76.1} & \textbf{94.0} & \textbf{57.6} & \textbf{89.2} & \textbf{50.2} & \textbf{75.8} \\
$\mathcal{N}$=0.40 & \textbf{88.1} & \textbf{98.8} & \textbf{76.7} & \textbf{94.0} & 56.7 & 88.9 & 47.8 & 75.4 \\
$\mathcal{N}$=0.45 & 87.8 & 98.6 & 75.9 & 93.8 & 53.5 & 88.4 & 46.1 & 74.3 \\
$\mathcal{N}$=0.50 & 87.8 & 98.4 & 74.8 & 93.2 & 53.2 & 88.1 & 45.3 & 74.1 \\
\bottomrule
\end{tabular}
\end{table}

\section{Conclusion}
In this study, we introduce a novel point cloud matching model called the Saliency-guided Transformer (D3Former). This model is designed for the joint learning of repeatable dense detectors and feature-enhanced descriptors, incorporating the FEDL module and RKDL module. Our proposed method excels in detecting repeatable keypoints, leading to a substantial enhancement in feature discriminability within texture-missing regions and an improved matching capability in regions with rich textures.
Extensive experiments conducted on indoor and outdoor benchmarks showcase the superior performance and robustness of our proposed method.




\bibliography{reference}

\end{document}